# A Right-of-Way Based Strategy to Implement Safe and Efficient Driving at Non-Signalized Intersections for Automated Vehicles

Yadong Xing, Can Zhao, Zhiheng Li, *Member, IEEE*, Yi Zhang, *Member, IEEE*, Li Li, *Fellow, IEEE*, Fei-Yue Wang, *Fellow, IEEE*, Xiao Wang, Yujing Wang, Yuelong Su, and Dongpu Cao, *Member, IEEE*

*Abstract*—Non-signalized intersection is a typical and common scenario for connected and automated vehicles (CAVs). How to balance safety and efficiency remains difficult for researchers. To improve the original Responsibility Sensitive Safety (RSS) driving strategy on the non-signalized intersection, we propose a new strategy in this paper, based on right-of-way assignment (RWA). The performances of RSS strategy, cooperative driving strategy, and RWA based strategy are tested and compared. Testing results indicate that our strategy yields better traffic efficiency than RSS strategy, but not satisfying as the cooperative driving strategy due to the limited range of communication and the lack of long-term planning. However, our new strategy requires much fewer communication costs among vehicles.

## I. INTRODUCTION

The technique of connected and automated vehicles (CAVs) provides shorter car-following gaps and faster responses in most traffic scenarios [1]-[2]. Many researchers regard it as a promising method to reduce traffic congestion and improve traffic efficiency. However, how to balance safety and efficiency remains difficult for researchers, because these two do not coincide with each other in many situations.

To ensure safety, researchers of Mobileye proposed a mathematical model for automated vehicles, which is called Responsibility Sensitive Safety (RSS) [3]. RSS strategy aims to ensure the automated vehicles will cause no accidents and properly respond to the mistakes of other drivers. However, as pointed out and analyzed particularly in [4] and [5], the performance of the original RSS strategy is still not satisfying enough from the viewpoint of traffic efficiency. RSS strategy is considered as a conservative decision-making strategy based on the intelligence of a single vehicle [6]. Following the same idea of RSS strategy to ensure absolute safety, some new strategies were proposed in [4] and [5] respectively to improve the efficiency performance on car following and lane change scenarios. These new approaches addressed the importance of communication between vehicles and highlight the transfer of the right-of-way in the scenarios.

As discussed in [4] and [5], we focus on another typical scenario in this paper, which is the non-signalized intersection. Collisions around intersections contribute to a significant portion of highway accidents and have aroused widespread concern in academic circles in recent years [7]-[17].

For the non-signalized intersection, RSS applies the typical First-In-First-Out (FIFO) strategy with few communications [3]. It regards there is no need for vehicles to interact, which means drive intentions of other vehicles cannot be gotten in advance. In order to ensure absolute safety in this scenario, an important rule of RSS is "right-of-way is given, not taken" [3], which means the vehicle must always be ready to brake and stop in case of a possible collision. However, this simple rule will lead to widespread congestion at the intersection in many cases, which has been confirmed by our experiments.

Cooperative driving is another common strategy of non-signalized intersections. Its major idea is applied Vehicle-to-everything (V2X) communication to organize and coordinate the movement of neighboring vehicles to get a long-term planning [7]-[12]. Passing order is the major factor of the traffic efficiency because the time points of vehicles passing through the potential conflicting points are determined by it [1]. Therefore, cooperative driving generally improves the overall traffic efficiency by optimizing the passing order. In this paper, we choose the Monte Carlo tree search (MCTS) strategy as a representative implementation of cooperative driving strategy [12]. Because it can find a nearly global-optimal passing order within a short planning time.

Since appropriate interaction and communication improve traffic efficiency in specific scenarios [4]-[5], [12], we propose a new strategy based on right-of-way assignment (RWA) to improve the RSS strategy. However, unlike the long-term cooperative trajectory planning as MCTS, our strategy roughly follows FIFO, but shorten the safety gap via appropriate right-of-way assignment.

To intuitively compare the traffic efficiency of the original RSS strategy, RWA based strategy, and the cooperative driving strategy respectively, we record the average delay and traffic throughput through simulation experiments. The testing results show that the new strategy can keep a balance of safety and traffic efficiency.

Manuscript received April 18th 2019. This work was supported in part by the National Natural Science Foundation of China (61790565), Science and Technology Innovation Committee of Shenzhen (JCYJ20170818092931604), the Beijing Municipal Commission of Transport Program (ZC179074Z), Intel Collaborative Research Institute for Intelligent and Automated Connected Vehicles (ICRI-IACV), and Joint Laboratory for Future Transport and Urban Computing of AutoNavi. (Corresponding author: *Li Li*)

Y. Xing, C. Zhao, Z. Li, and Y. Zhang are with the Department of Automation, Tsinghua University, Beijing 100084, China. Z. Li is also with the Graduate School at Shenzhen, Tsinghua University, Shenzhen 518055,China.

L. Li is with Department of Automation, BNRist, Tsinghua University, Beijing 100084, China (email: li-li@tsinghua.edu.cn).

F.-Y. Wang and X. Wang are with the State Key Laboratory for Management and Control of Complex Systems, Institute of Automation, Chinese Academy of Sciences, Beijing 100190, China.

Y. Wang and Y. Su are with traffic management solution division, AutoNavi Software Co, Beijing 100102, China.

D. P. Cao is with the Driver Cognition and Automated Driving Laboratory, University of Waterloo, Waterloo N2L 3G1, Canada.

To give a better presentation of our findings, the rest of this paper is arranged as follows. *Section II* presents the problem and analyzes three different strategies separately. *Section III* presents our RWA based strategy in detail. *Section IV* compares RWA based strategy with the original RSS strategy and the cooperative driving strategy and provides testing results. Finally, *Section V* concludes the paper.

II. PROBLEM PRESENTATION AND STRATEGY ANALYSIS

Because of the cost and some other reasons, many intersections on the country roads, military zones or industrial zones have no traffic lights [8]. In this paper, we focus on vehicle driving scenarios at such non-signalized intersections, as studied in [8] and [10].

*A. Problem Presentation*

Fig.1 shows a typical non-signalized intersections scenario with four branches and eight lanes. The lanes are labeled clockwise. The area within the circle is called *the control zone* and the shadow area is called *the junction* where lateral collisions might happen. Vehicles can go straight, turn left, and turn right at the intersection. There is an example of four vehicles passing through *the junction* as Fig.1 shows. **Vehicle A**, **B**, **C**, and **D** are the focus of our study. Because conflicts between different vehicles at intersections can be resolved in the same way, we choose **A** as the studied vehicle.

In addition, there is a precondition that each vehicle follows the same strategy to determine the right-of-way when potential conflicts/collisions are detected. We assume that lane change is not allowed within *the control zone* since there are only little gaps between vehicles and one lane per direction.

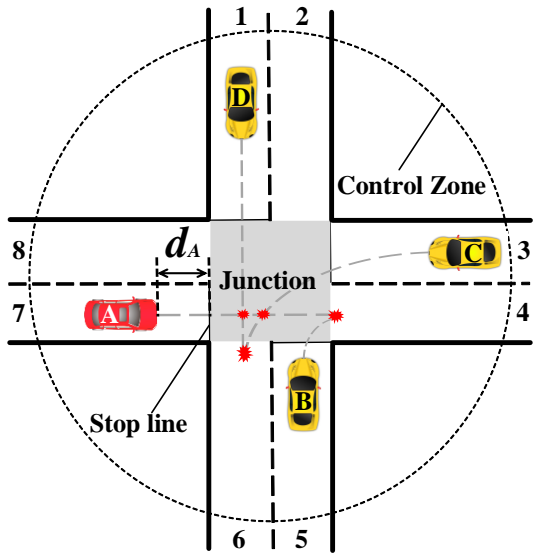

**Fig.1** An illustration of the non-signalized intersection [10].

In this scenario, collisions can only happen in longitudinal for car following or in lateral for intersection passing. We adopt the car following strategy of [4] and just focus on the latter in this paper. According to the trajectory intersect of vehicles from different direction，collisions at *the junction* mainly occur in the potential conflicting points, so can be summarized into seven cases, as shown in Fig.2. Rotation invariance has been taken into account. Besides the locations of collisions, we also consider drive intentions, so Fig.2 (a), (b) and (c) are considered as three different scenarios. When the studied vehicle can interact with other vehicles on by one, we regard the problem of the multi-vehicle conflict in Fig.1 can be decomposed into four two-vehicle conflicts to simplify the problem. We can see that all four potential conflicts in Fig.1 can be found in Fig.2.

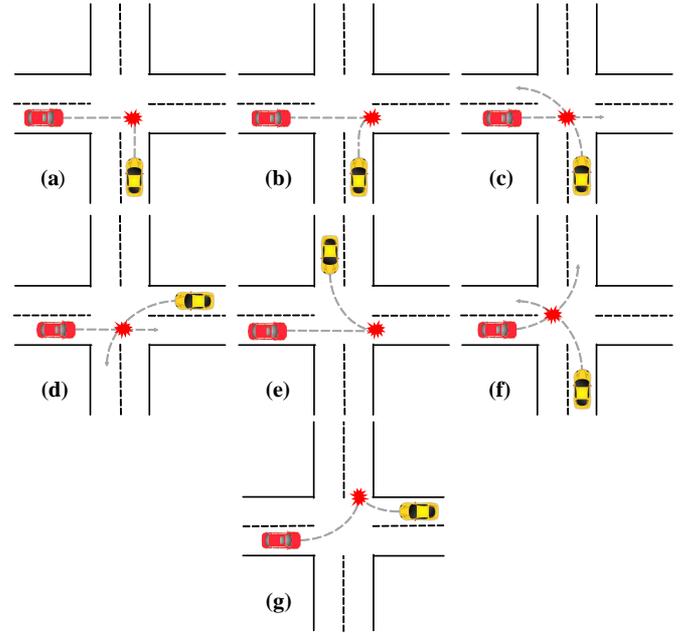

**Fig.2** All 7 possible collision scenarios of two vehicles at the junction.

We regard the key idea to solve these collisions/conflicts is to clarify the ownership of right-of-way at the junction. So that the owner of the right-of-way can get priority naturally and the other should take responsibility to keep a safety gap. A clear right-of-way assignment strategy will guide vehicles to pass the intersection with both high efficiency and safety.

*B. Analysis of Three Strategies*

The biggest difference between the three strategies is regarded the amount of communication. From the RSS strategy to the cooperative driving strategy, information of vehicles and the required interactions between vehicles are gradually increasing. Passing orders are the outcome of three strategies in different methods. In some cases, these three strategies may get the same passing order.

1) *The RSS Strategy*

For ensuring absolute safety, the RSS strategy adopts the following assumptions:

- The studied vehicle can only get the position and speed information of other vehicles.

- Only one vehicle pass through *the junction* at a time since the drive intention of others is unavailable.

According to the three scenes that the vehicle will encounter at the intersection, the intersection passing process can be divided into three stages, *the car-following stage*, *the decision stage*, and *the action stage*. The flow chart of the RSS strategy as shown in Fig.3 (a). It determines the priority of vehicles passing through *the junction* via longitudinal distance, e.g. $d_A$ in Fig.1 is the longitudinal distance of **A**.

In *the decision stage*, if **A** is the closest vehicle, it will pass *the junction* in the highest priority. If not, it should wait in the stop line, then passing *the junction* according to the FIFO.

It is a conservative strategy leads to low traffic efficiency, since it assumes *the junction* can only be occupied by one vehicle at one moment. However, two vehicles can pass *the junction* without any conflict. The RSS strategy cannot solve the possible "deadlock" problem, e.g. four vehicles have the same longitudinal distance.

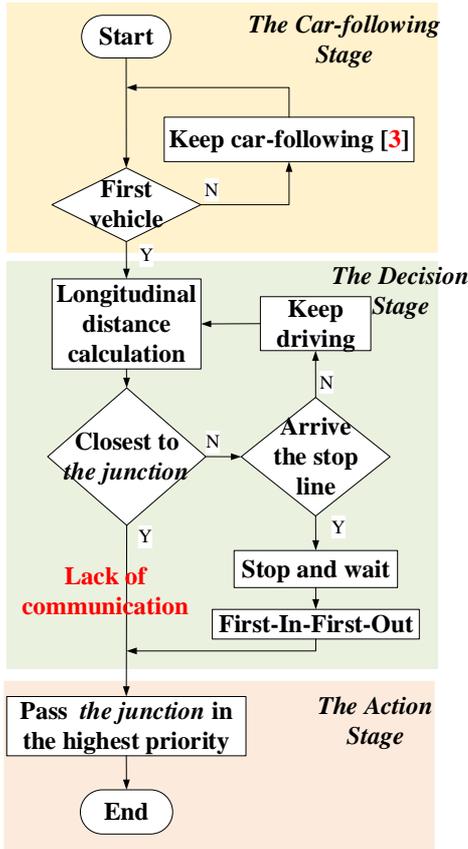

(a) The RSS strategy [3]

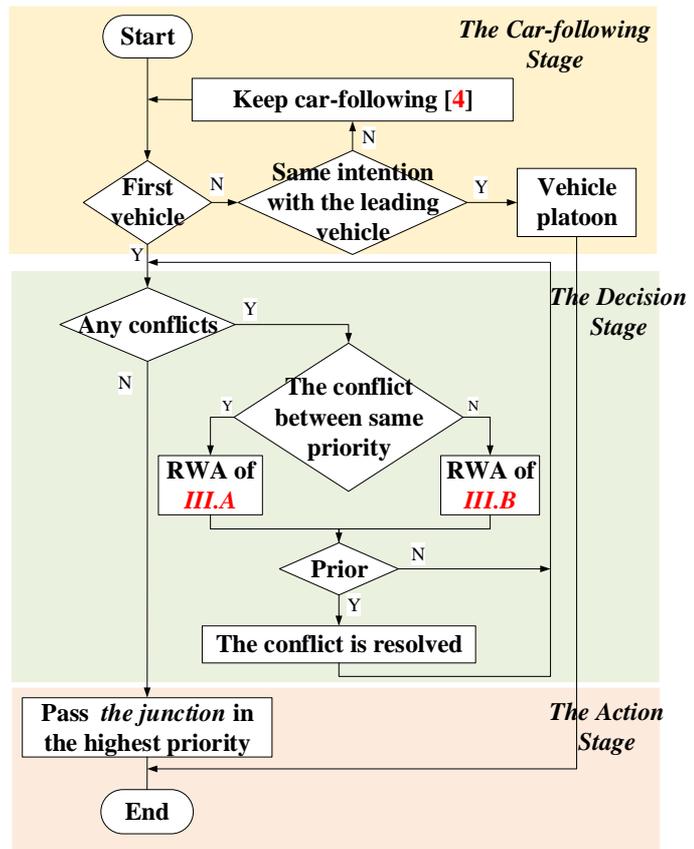

(b) The Right-of-way assignment (RWA) based strategy.

**Fig.3** The flow chart of the RSS strategy and the RWA based strategy.

2) *The RWA based Strategy*

Based on the transfer of right-of-way among vehicles, we propose a new driving strategy as shown in Fig.3 (b). To avoid the potential "deadlock" and improve traffic efficiency, the RWA based strategy introduce more communication as well as clear rules of the right-of-way assignment.

For the right-of-way, the definition has been given as "**the right to occupy/use a special temporal-spatial area**" in [5] and [18]. In this paper, we also use this definition and the other four basic rules of [5] to determine the passing order via right-of-way assignment. Moreover, our strategy makes short-term decisions about the right-of-ways for certain temporal-spatial areas and reaches these decisions by using one-to-one communications between vehicles.

For simplicity, we adopt the following assumptions:

- Allow two vehicles to pass the junction at the same time.
- The studied vehicle can get the position, speed and drive intention information of other vehicles.
- General priority level is defined as straight-going vehicles > left-turning vehicles > right-turning vehicles.
- Among vehicles of the same priority level (e.g. two straight-going vehicles from different directions), the closer one (to trajectory overlap point) is prior.

- When the other conditions are the same, the vehicles on lanes 3 and 7 take precedence over lanes 1 and 5.

In *the car-following stage*, our strategy applies the car following model in [4], because it is more reasonable. Besides, consecutive vehicles with same drive intention will form a vehicle platoon [8]. These vehicles will be treated as a group, passing *the junction* in sequence.

In *the decision stage*, the first vehicle will communicate with the vehicles at *the junction* and the first car of other three lanes one by one. If there is a potential conflict, the strategy can resolve it by a clear assignment of the right-of-way, which will be analyzed in detail in *section III*. So the problem of multi-vehicle passing through the intersection can be decomposed into several independent right-of-way assignment cases between two vehicles.

In *the action stage*, all the conflicts of the studied vehicle are resolved, it will get the right-of-way of *the junction* and safely passing.

3) *The Cooperative Driving Strategy*

The cooperative driving strategy assumes that the information about all traffic participants (vehicles, pedestrians, etc.) is available via V2X communication. The V2X system can cover a range of up to 300m or more [19].

It solves the problem from another respect, regarding the problem as a global optimization problem. It takes the vehicles within a wide-range area into account and aims to find a time-optimal passing order. The cooperative driving strategy divides *the junction* into several separate subzones with gridding method and then sets the minimum time interval between vehicles passing the same subzone as a constant (e.g. 2 seconds in MCTS) to avoid potential conflicts. However, more communications and computation resources will be spent inevitably during this process, which means the results will be very likely not available in a limited time as the number of computation increases.

Cooperative driving is not the focus of this paper, hence omitted the details, which can be found in [12].

For the four vehicles in the scenario of Fig.1, the passing order of the three strategies may be different. The passing order is **B**, **A**, **D**, **C** for the RSS strategy, **A, (B, D), C** for the RWA based strategy. The vehicles in parentheses can pass the junction at the same time. For the cooperative driving, the passing order varies with the number of vehicles calculated, (**B, C), A, D** is one possible order of it.

III. RIGHT-OF-WAY ASSIGNMENT BETWEEN TWO VEHICLES

In this section, we discuss how to assign the right-of-way between two vehicles from different branches, and 7 possible conflicts listed above are guaranteed to avoid.

As a representative, we chose the scenario of Fig.2 (a) and Fig.2 (b). Because those two scenarios represent vehicles with the same drive intention and the different drive intentions respectively. The other five scenarios can be dealt with by the approximate method.

In order to assign the right-of-way among vehicles from different lanes properly, we introduce virtual vehicle mapping technique into this part [20]-[21]. By means of this technique, two vehicles in different lanes can negotiate and adjust the relative location as located in the same lane.

In *the control zone*, low priority vehicle will be mapped to the lane of higher priority vehicle when there is a potential collision. The road around the high priority vehicle can be divided into *The Forbidden Area*, *The Negotiation Area*, and *The Free Area* which is defined in [5], as Fig.4 and Fig.5 show. *The Forbidden Area* means no vehicles are allowed to enter, and *The Negotiation Area* means that other vehicles have a chance to obtain the right-of-way of this area after negotiating with the owner.

*A. The RWA of Two Vehicles in the Same Drive Intention*

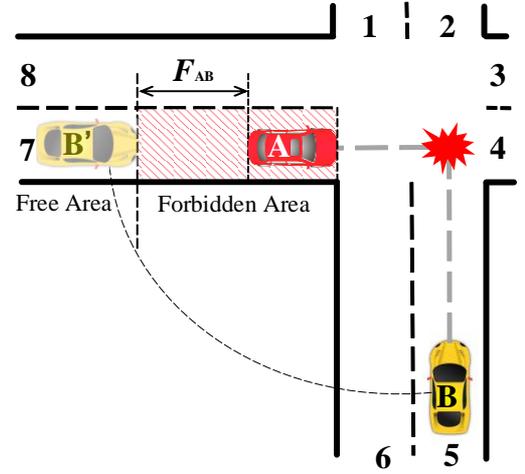

**Fig.4** Right-of-way assignment between two vehicles in same drive intention.

For the vehicles as Fig.2 (a) and (f) show, the closer one (to trajectory overlap point) is prior. As Fig.4 shows, **Vehicle A** is closer, so **A** is prior and **B'** (virtual vehicle mapped from **B**) cannot drive into *the forbidden area* behind **A**. The length of *The Forbidden Area* is defined as

$$F_{AB} = \left[ v_B \rho + \frac{v_B^2}{2b_{max,brake}} - \frac{v_A^2}{2a_{max,brake}} \right]^+ \quad (1)$$

Where $v_A$ and $v_B$ is the speed of **A** and **B**, $a_{max,brake}$ and $b_{max,brake}$ is the max braking deceleration of them, $\rho$ is the response time of the driver.

*B. The RWA of Two Vehicles in Different Drive Intentions*

For the vehicles in the different drive intentions as Fig.2 (b), (c), (d), (e) and (g) show, we follow the priority level which is defined in *Section II*. As Fig.5 shows, straight-going **Vehicle A** owns the absolute right-of-way of *The Forbidden Area* and the relative right-of-way of *The Negotiation Area*. **B'** is a virtual vehicle mapped from **B**.

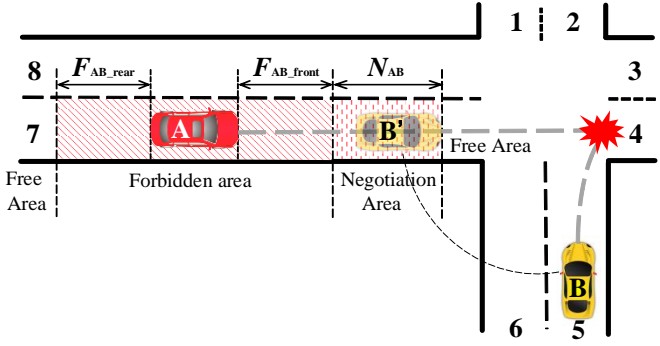

**Fig. 5** Right-of-way assignment between two vehicles in different intentions.

According to the location of **Vehicle B'**, there are three cases for the right-of-way assignment between **A** and **B**:

1) If **B'** is in *The Forbidden Area* ahead or behind of **A**, then a collision will happen in all probability. So **B** should brake and stop until **B'** reach *The Free Area* behind **A**.

2) If **B'** is in *The Negotiation Area* of **A**, **A** owns right-of-way and **B** can negotiate with **A**. **A** can accept or reject the request. The right-of-way will transfer to **B** if **A** accepted the request. So **B** can go on turning right, and **A** should keep a safety gap from **B'**. On the contrary, **B** should brake and stop at once until **B'** reach *The Free Area* behind **A**.

3) If **B'** is in *The Free Area* of **A**, the collision will never happen obviously. In this scenario, both **A** and **B** can keep the original state of motion to pass the intersection safely.

The length of *The Forbidden Area* behind **A** can be calculated as

$$F_{AB\_rear} = \left[\frac{v_B^2}{2b_{min,brake}} - \frac{v_A^2}{2a_{max,brake}}\right]^+ \quad (2)$$

Where $b_{min,brake}$ is the minimum deceleration rate of **B**. Because the braking rate of real vehicles cannot be approximated to zero continuously, it is a discrete physical system.

The distance between the front point of **A** and the boundary of *The Free Area* is

$$F_{AB\_front} + N_{AB} = \left[v_A \rho + \frac{v_A^2}{2a_{min,brake}} - \frac{v_B^2}{2b_{max,brake}}\right]^+ \quad (3)$$

Where $a_{min,brake}$ is the minimum deceleration rate of **A**.

After the ownership of the right-of-way is clear and the potential conflicts are resolved, the vehicle can pass through *the junction* safely in the proper order.

## IV. SIMULATION TESTING RESULTS

In order to evaluate the performance of RWA based strategy and RSS strategy, we designed the following experiments and compared with the cooperative driving strategy.

### A. Simulation Settings

The vehicles' arrival rate (traffic demand) at each branch is assumed to be a Poisson Process whose average value is denoted as λ [10]. Varying the values of λ, we can check the performance of each strategy on different traffic flows. For each branch, the ratio of left-turning, right-turning, and straight-going vehicles is 3:3:4, and they are randomly distributed. Vehicles are generated 200 meters away from the center of *the junction*. We assume that the owner of the right-of-way will accept the request in a half probability. The point-queue model is applied in this paper to accurately calculate the total delays of vehicles [10].

Particularly, we set $a_{max,brake} = b_{max,brake} = 6 \text{m/s}^2$, $a_{min,brake} = b_{min,brake} = 2 \text{m/s}^2$ and $\rho = 1\text{s}$ in the rest tests. The radius of *the control zone* is set as 100m, which is alterable according to the sensors and other conditions. The width of the lane is set as 3m, and the vehicle length is set as 3.5m. The limited speed for straight-going vehicles is 10m/s and for left/right-turning vehicles is 5m/s. We set the simulation interval as 0.1s and the throughput is calculated within 20 minutes.

### B. The Results of the Average Delay and Throughput

We record the average delay and throughput [14] to evaluate different strategies, whose results are shown in Fig.6 and Fig.7.

From the results, we can observe huge differences in the performance of three strategies. As shown in Fig.6, the results of the RSS strategy has a much larger delay time than the RWA based strategy and cooperative driving strategy (MCTS), especially when the arrival rate becomes large. The average delay of the RSS strategy still arrives 20 seconds even we set the traffic demand as 200 veh/(lane·h), which means this strategy is likely to lead to congestion in most cases.

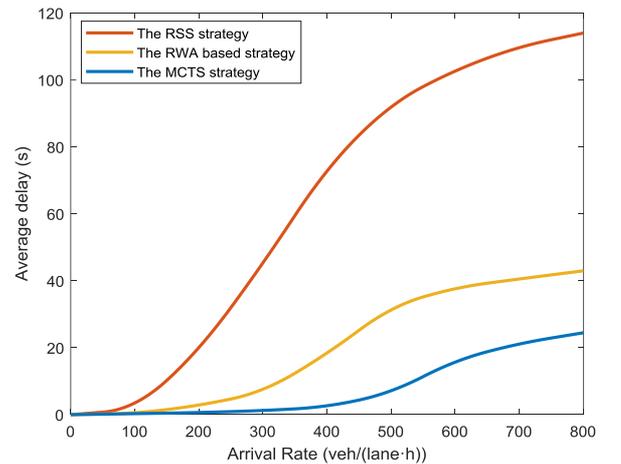

**Fig.6** The average delay of three strategies.

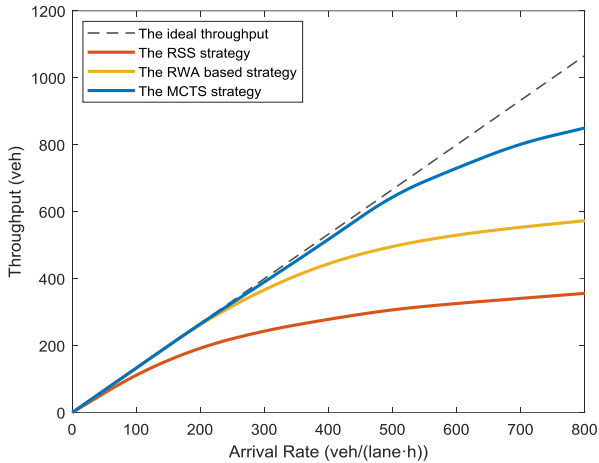

**Fig.7** The throughput of three strategies compared with the ideal throughput.

Fig.7 shows the throughput of the intersection obtained from different strategies in 20 minutes and compared with the ideal result. As the flow rate increase from 400 to 800, the throughput is almost the same for the RSS strategy, which means the intersection is already blocked. As Fig.6 and Fig.7 show, we can observe that the congestion has already formed when the arrival rate get 200 veh/(lane·h) for RSS strategy. The number is 400 for the RWA based strategy, and 600 for the cooperative driving strategy.

The RWA based strategy yields better traffic efficiency than RSS strategy due to the communication and clear assignment of the right-of-way. Although the cooperative driving strategy can get better traffic efficiency, it reaches the performance at the cost of more information transfer and computation. As analyzed in [10], the number of passing orders and the computational time grows almost exponentially with respect to the number of vehicles. The RWA based strategy overcomes this problem well through distributed computing of every single vehicle. From this perspective, the RWA based strategy achieves a practical level of efficiency with less information interaction and less computation, which means it achieves a balance between traffic efficiency and usability.

## V. Conclusion

In this paper, we propose a driving strategy based on the right-of-way for non-signalized intersection, aiming to find a balance of collision-free and traffic efficiency. We simplify the problem with the aid of short-term communication and virtual vehicle mapping technique. Testing results indicate that the performance of the proposed strategy is much better than RSS.

The RWA based strategy uses far less communication and computing resources but achieves performance comparable to the cooperative driving strategy. Therefore, we believe that it is a scalable strategy applied for the automated vehicle, which has great research value and development prospects in the future.